\documentclass[10pt, a4paper]{article}
\usepackage{lrec}
\usepackage{booktabs}
\usepackage{multirow}

\title{GitHub Typo Corpus: \\ A Large-Scale Multilingual Dataset of Misspellings and Grammatical Errors}

\name{Masato Hagiwara\textsuperscript{1} and
      Masato Mita\textsuperscript{2, 3}}

\address{\textsuperscript{1}Octanove Labs, Seattle, WA, USA \\
         \textsuperscript{2}RIKEN AIP, Tokyo, Japan \\
         \textsuperscript{3}Tohoku University, Miyagi, Japan \\
         \textsuperscript{1}masato@octanove.com, 
         \textsuperscript{2}masato.mita@riken.jp}

\abstract{
The lack of large-scale datasets has been a major hindrance to the development of NLP tasks such as spelling correction and grammatical error correction (GEC). As a complementary new resource for these tasks, we present the GitHub Typo Corpus, a large-scale, multilingual dataset of misspellings and grammatical errors along with their corrections harvested from GitHub, a large and popular platform for hosting and sharing git repositories. The dataset, which we have made publicly available, contains more than 350k edits and 65M characters in more than 15 languages, making it the largest dataset of misspellings to date. We also describe our process for filtering true typo edits based on learned classifiers on a small annotated subset, and demonstrate that typo edits can be identified with $F1 \sim 0.9$ using a very simple classifier with only three features. The detailed analyses of the dataset show that existing spelling correctors merely achieve an F-measure of approx. 0.5, suggesting that the dataset serves as a new, rich source of spelling errors that complement existing datasets. \\ \newline  }

\begin{document}

\maketitleabstract

\section{Introduction}

Spelling correction \cite{Islam:2009,Zhou:2017,Etoori:2018} and grammatical error correction (GEC) \cite{Leacock:2010} are two fundamental tasks that have important implications for downstream NLP tasks and for education in general. In recent years, the use of statistical machine translation (SMT) and neural sequence-to-sequence (seq2seq) models has been becoming increasingly popular for solving these tasks. Such modern NLP models are usually data hungry and require a large amount of parallel training data consisting of sentences before and after the correction. However, only relatively small datasets are available for these tasks, compared to other NLP tasks such as machine translation. This is especially the case for spelling correction, for which only a small number of datasets consisting of individual misspelled words are available, including the Birkbeck spelling error corpus\footnote{\url{http://hdl.handle.net/20.500.12024/0643}} and a list of typos collected from Twitter\footnote{\url{http://luululu.com/tweet/}}.

Due to this lack of large-scale datasets, many research studies \cite{Foster:2009,Etoori:2018,Li:2018} resort to automatic generation of artificial errors (also called pseudo-errors). Although such methods are efficient and have seen some success, they do not guarantee that generated errors reflect the range and the distribution of true errors made by humans \cite{Zesch:2012}.

\begin{figure}[!t]
\begin{center}
\includegraphics[scale=0.43]{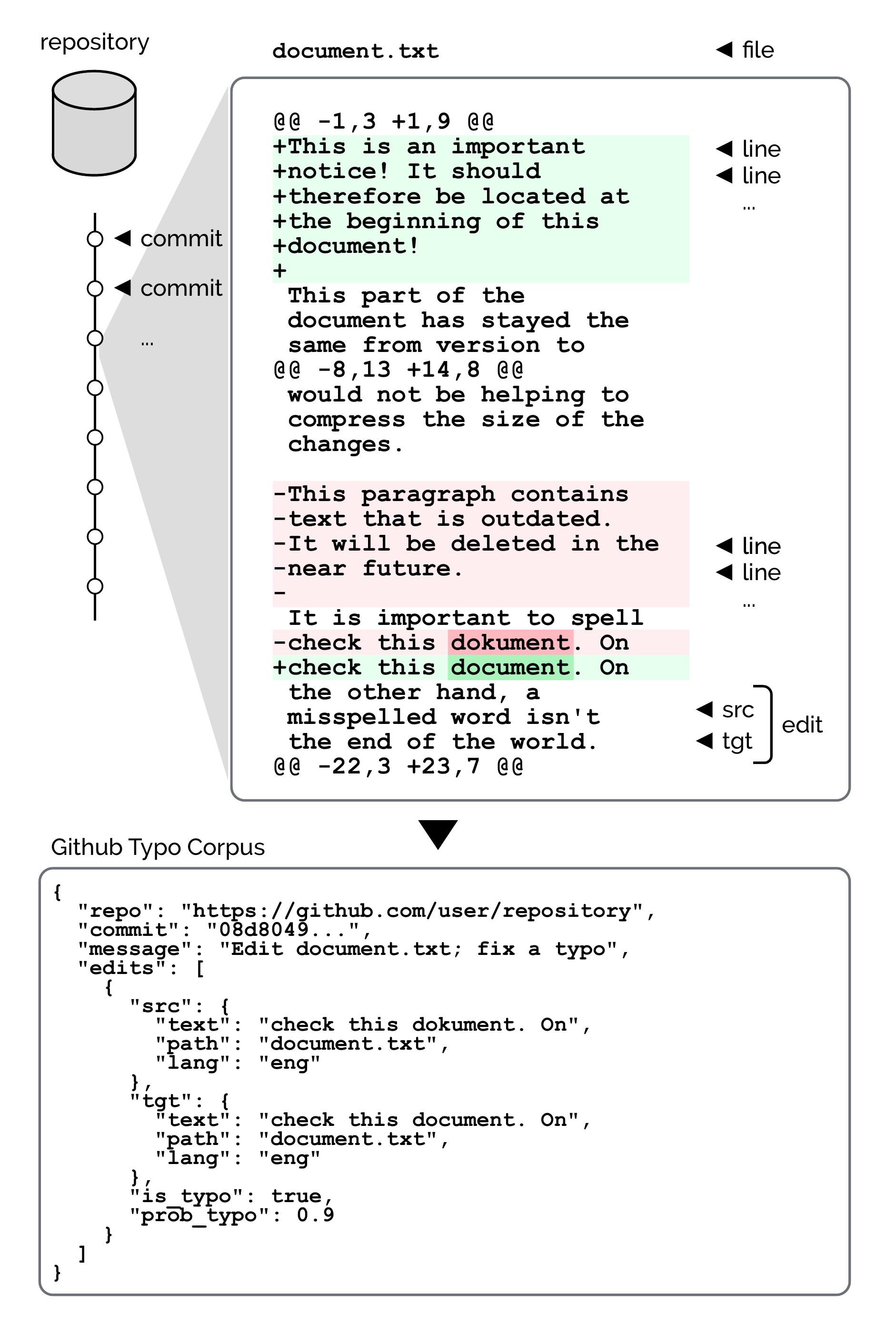} 
\caption{Overview of the corpus and its related concepts.\\
         Example taken from the Diff page on Wikipedia}
\label{fig:overview}
\end{center}
\end{figure}

As one way to complement this lack of resources, Wikipedia has been utilized as a rich source of textual edits, including typos \cite{Grundkiewicz:2014,Boyd:2018,Faruqui:2018}. However, the edits harvested from Wikipedia are often very noisy and diverse in their types, containing edits from typos to adding and modifying information. To make the matters worse, Wikipedia suffers from vandalism, where articles are edited in a malicious manner, which requires extensive detection and filtering. 

In order to create a high-quality, large-scale dataset of misspelling and grammatical errors (collectively called {\it typos} in this paper), we leverage the data from GitHub\footnote{\url{https://github.com/}}, the largest platform for hosting and sharing repositories maintained by git, a popular version control system commonly used for software development. Changes made to git repositories (called {\it commits}, see Section 3 for the definition) are usually tagged with commit messages, making detection of typos a trivial task. Also, GitHub suffers less from vandalism, since commits in many repositories are code reviewed, a process where every change is manually reviewed by other team members before merged into the repository. This guarantees that the edits indeed fix existing spelling and/or grammatical issues.

This paper describes our process for building the GitHub Typo Corpus, a large-scale, multilingual dataset of misspellings and grammatical errors, along with their corrections. The process for building the dataset can be summarized as follows:
\begin{itemize}
    \item Extract eligible repositories and typo commits from GitHub based on the meta data of the repository and the commit message
    \item Filter out edits that are not written in human language
    \item Identify true typo edits (vs semantic edits) by using learned classifiers on a small annotated dataset
\end{itemize}

We demonstrate that a very simple logistic regression model with only three features can classify typos and non-typo edits correctly with $F1 \sim 0.9$. This resulted in a dataset containing more than 350k edits and 64M characters in more than 15 languages. To the best of our knowledge, this is the largest multilingual dataset of misspellings to date. We made the dataset publicly available (\url{https://github.com/mhagiwara/github-typo-corpus}) along with the automatically assigned typo labels as well as the source code to extract typos. We also provide the detailed analyses of the dataset, where we demonstrate that the F measure of existing spell checkers merely reaches $\sim 0.5$, arguing that the GitHub Typo Corpus provides a new, rich source of naturally-occurring misspellings and grammatical errors that complement existing datasets.

\section{Related Work}

As mentioned above, a closely related line of work is the use of Wikipedia edits for various tasks, including GEC. \newcite{Grundkiewicz:2014} constructed the WikiEd Error Corpus, a dataset consisting of error edits harvested from the Wikipedia edit history and demonstrated that the newly-built resource was effective for improving the performance of GEC systems. \newcite{Boyd:2018} built a German GEC system leveraging the WikiEd Error Corpus and showed that the use of the Wikipedia edit data led to improved performance. In both cases, the dataset required extensive filtering based on a set of heuristic rules or heavy linguistic analysis.

Spelling correction is itself an important sub-problem of grammatical error correction (GEC). Many GEC and essay scoring systems \cite{Sakaguchi:2017,Junczys-Dowmunt:2018,Vajjala:2018} assume that spelling errors in the input text are fixed before it is fed to the main model, by pre-processing them using open-source tools such as Enchant\footnote{\url{https://github.com/AbiWord/enchant}} and LanguageTool\footnote{\url{https://languagetool.org/}}. In many GEC corpora, spelling errors account for approximately 10\% of total errors (Table \ref{table:spelling_errors}), meaning that improving the accuracy of spelling correction can have a non-negligible impact on the performance of GEC.

\begin{table}[!ht]
\begin{center}
\begin{tabular}{@{}lr@{}}
\toprule
Corpus & Misspellings (\%)  \\ \midrule
CLC-FCE \cite{Yannakoudakis:2011} & 9.69 \\
JFLEG \cite{Napoles:2017}         & 12.56 \\
KJ \cite{Nagata:2011}             & 9.41 \\ \bottomrule
\end{tabular}
\caption{Percentage of spelling errors in GEC corpora}
\label{table:spelling_errors}
\end{center}
\end{table}

Datasets of real-world typos have applications in building models robust to spelling errors \cite{Piktus:2019}. We note that \newcite{Mizumoto:2017} argue against the necessity of spell checking on learner English, which has little effect on the performance of PoS (part-of-speech) tagging and chunking.

\section{Definitions}

First, we define and clarify the terminology that we use throughout this paper. See Figure \ref{fig:overview} for an illustration of the concepts and how they relate to each other.

\begin{itemize}
    \item Repository ... in git terms, a repository is a database of files whose versions are controlled under git. A single repository may contain multiple files and directories just like a computer file system.
    \item Commit ... a commit is a collection of one or more changes made to a git repository at a time. Changes in a single commit can span over multiple files and multiple parts of a file.
    \item Edit ... in this paper, an edit is a pair of lines to which changes are made in a commit (note the special usage here). The line before the change is called the {\it source} and the line after is the {\it target}. In other words, an edit is a pair of the source and the target. Note that a single edit may contain changes to multiple parts of the source (for example, multiple words that are not contiguous).
    \item Typo ... finally, in this paper a typo refers to an edit where the target fixes some mechanical, spelling and/or grammatical errors in the source, while preserving the meaning between the two.
\end{itemize}

Our goal is to collect typos from GitHub and build a dataset that is  high in both quantity and quality.

\section{Data Collection}
\begin{figure}[!t]
\begin{center}
\includegraphics[scale=0.5]{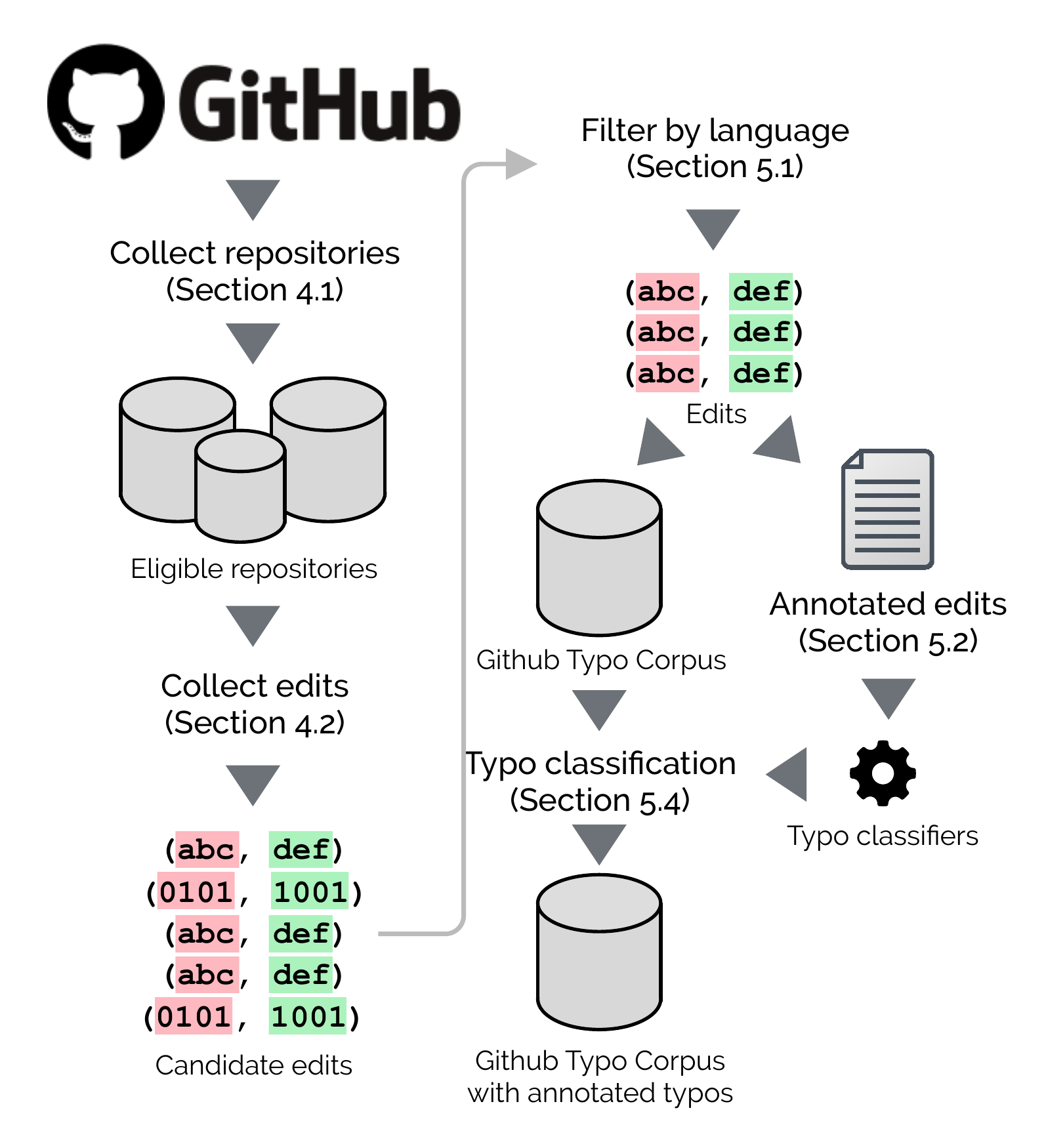}
\caption{Data collection and filtering process}
\label{fig:process}
\end{center}
\end{figure}

This section describes the process for collecting a large amount of typos from GitHub, which consists two steps: 1) collecting target repositories that meet some criteria and 2) collecting commits and edits from them. See Figure \ref{fig:process} for the overview of the typo-collecting process.

\subsection{Collecting Repositories}

The first step for collecting typos is to collect as many eligible GitHub repositories as possible from which commits and edits are extracted. A repository must meet some criteria in order to be included in the corpus, such as size (it needs to big enough to contain at least some amount of typo edits), license (it has to be distributed under a permissive license to allow derived work), and quality (it has to demonstrate some signs of quality, such as the number of stars).

Although GitHub provides a set of APIs (application programming interfaces) that allow end-users to access its data in a programmatic manner, it doesn't allow flexible querying on the repository meta data necessary for our data collection purposes. Therefore, we turn to GH Archive\footnote{\url{https://www.gharchive.org/}}, which collects all the GitHub event data and make them accessible through flexible APIs. Specifically, we collected every repository from GH Archive that:

\begin{itemize}
    \item Has at least one pull request or pull request review comment event between November 2017 and September 2019,
    \item Has 50 or more starts,
    \item Has a size between 1MB and 1GB, and
    \item Has a permissive license.
\end{itemize}

Note the ``and'' in the list above—a repository needs to meet all the conditions mentioned above to be eligible. The first two criteria (pull request events and the number of starts) are a sign of a quality repository. As for the license, we allowed {\tt apache-2.0} (Apache License 2.0), {\tt mit} (MIT License), {\tt bsd-3-clause} (BSD 3-Clause License), {\tt bsd-2-clause} (BSD 2-Clause License), {\tt cc0-1.0} (Creative Commons Zero v1.0 Universal), {\tt unlicense} (Unlicense), {\tt cc-by-4.0} (Creative Commons Attribution 4.0), and {\tt bsl-1.0} (Boost Software License 1.0 (BSL-1.0). A repository's number of stars, size, and license are determined as of the event in the first condition.

This resulted in a total of 43,462 eligible repositories.

\subsection{Collecting Commits and Edits}

The second step for collecting typos is to extract commits and edits from the eligible repositories. This step is more straightforward—for each eligible repository, we cloned it using the GitPython library and enumerated all the commits in the master branch\footnote{For those who are not familiar with git, a branch is analogous to a ``version'' of a repository that you can create off of its main version, which is called the ``master'' branch.}. A commit is considered eligible if the commit message contains the string {\tt typo} in it. For each eligible commit, we then take the diff between the commit and its parent, scan the result sequentially, and collect all the pairs of a deletion line and a subsequent insertion line as an edit, unless the commit contains more than 10 edits, which is a sign of a non-typo commit. See the first box in Figure \ref{fig:overview} for an illustration. As a result, we collected a total of 335,488 commits and 685,377 edits. The final dataset (see the second box in Figure \ref{fig:overview} for a sample) is formatted in JSONL (JSON per line), where each line corresponds to a single commit with its metadata (its repository, commit hash, commit message, as well as a list of edits) in JSON, a format easily parsable by any programming language.

\section{Data Filtering}

Not all the edits collected in the process described so far are related to typos in natural language text. First, edits may also be made to parts of a repository that are written in programming language versus human language. Second, not every edit in a commit described ``typo'' is necessarily a typo edit, because a developer may make a single commit comprised of multiple edits, some of which may not be typo-related.

We remove the first type of edits by using language detection, and detect (not remove) the second type of edits by building a supervised classifier. The following subsections detail the process. See Figure \ref{fig:process} (right) for an overview of the typo filtering process.

\subsection{Language Detection}

Due to its nature, repositories on GitHub contain a large amount of code (in programming language) as well as natural language texts. We used NanigoNet\footnote{\url{https://github.com/mhagiwara/nanigonet}}, a language detector based on GCNNs (Gated Convolutional Neural Networks) \cite{Dauphin:2017} that supports human languages as well as programming languages. Specifically, we ran the language detector against both the source and the target and discarded all the edits where either is determined  as written in a non-human language. We also discarded an edit if the detected language doesn't match between the source and the target. This left us with a total of 203,270 commits and 353,055 edits, which are all included in the final dataset.

\subsection{Annotation of Edits}
\label{subsec:annotation}

\begin{figure*}[!t]
\begin{center}
\includegraphics[scale=0.5]{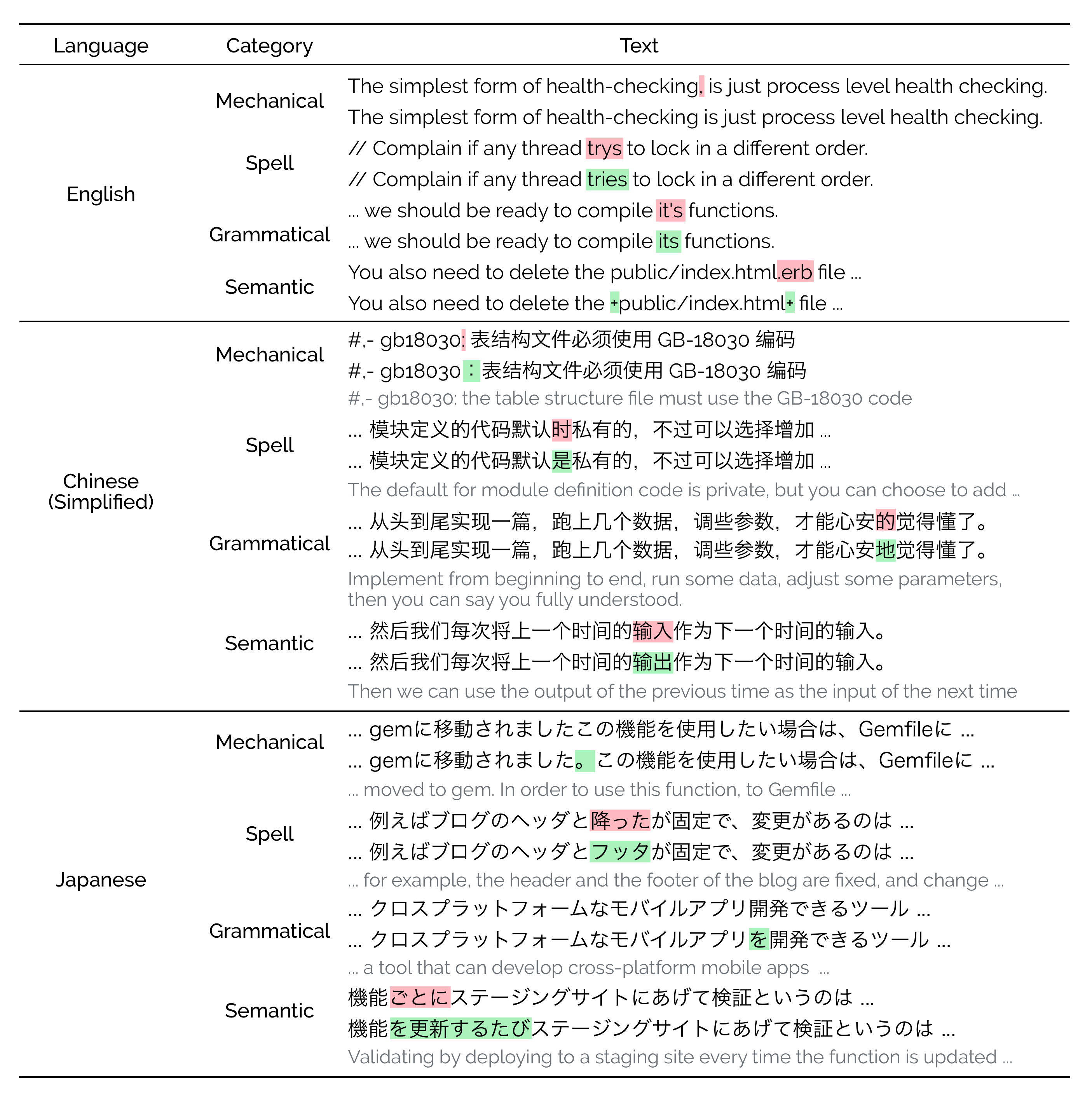}
\caption{Examples of different types of edits in top three languages}
\label{fig:examples}
\end{center}
\end{figure*}

In this second phase of filtering, we identify all non-typo edits that are not intended to fix mechanical, spelling, or grammatical errors, but to modify the intended meaning between the source and the target. 

In order to investigate the characteristics of such edits empirically, we first extracted 200 edits for each one of the three largest languages in the GitHub Typo Corpus: English ({\tt eng}), Simplified Chinese ({\tt cmn-hans}), and Japanese ({\tt jpn}). We then had fluent speakers of each language go over the list and annotate each edit with the following four edit categories:

\begin{itemize}
    \item Mechanical ... a mechanical edit fixes errors in punctuation and capitalization.
    \item Spell ... a spell edit fixes misspellings in words. This also includes conversion errors in non-Latin languages (e.g., Chinese and Japanese).
    \item Grammatical ... a grammatical edit fixes grammatical errors in the source.
    \item Semantic ... a semantic edit changes the intended meaning between the source and the target.
\end{itemize}

See Figure \ref{fig:examples} for some examples of different edit types on each language. If one edit contains more than one type of changes, the least superficial category is assigned. For example, if there are both spell and grammatical changes in a single edit, the ``grammatical'' category is assigned to the edit. We note that the first three (mechanical, spell, and grammatical edits, also called typos) are within the scope of the dataset we build, while the last one (semantic edits) is not. Thus, our goal is to identify the last type of edits as accurately as possible in a scalable manner. We will show the statistics of the annotated data in Section 6.

We note that the distinction between different categories, especially between spell and grammatical, is not always obvious. For example, even if one mistypes a word ``what'' to ``want'' resulting in an ungrammatical sentence, we wouldn't consider this as a grammatical edit but as a spell edit. We clarify the difference by focusing on the {\it process} where the error is introduced in the first place. Conceptually, if one assumes that the source is generated by introducing errors to the target through a noisy channel model \cite{Kernighan:1990,Brill:2000}, a spell edit is something where noise is introduced to some implicit character-generating process, while a grammatical edit is the one which corrupts some implicit grammatical process (for example, production rules of a context-free grammar).

\subsection{Statistics of Annotated Edits}

\begin{figure}[!t]
\begin{center}
\includegraphics[scale=0.5]{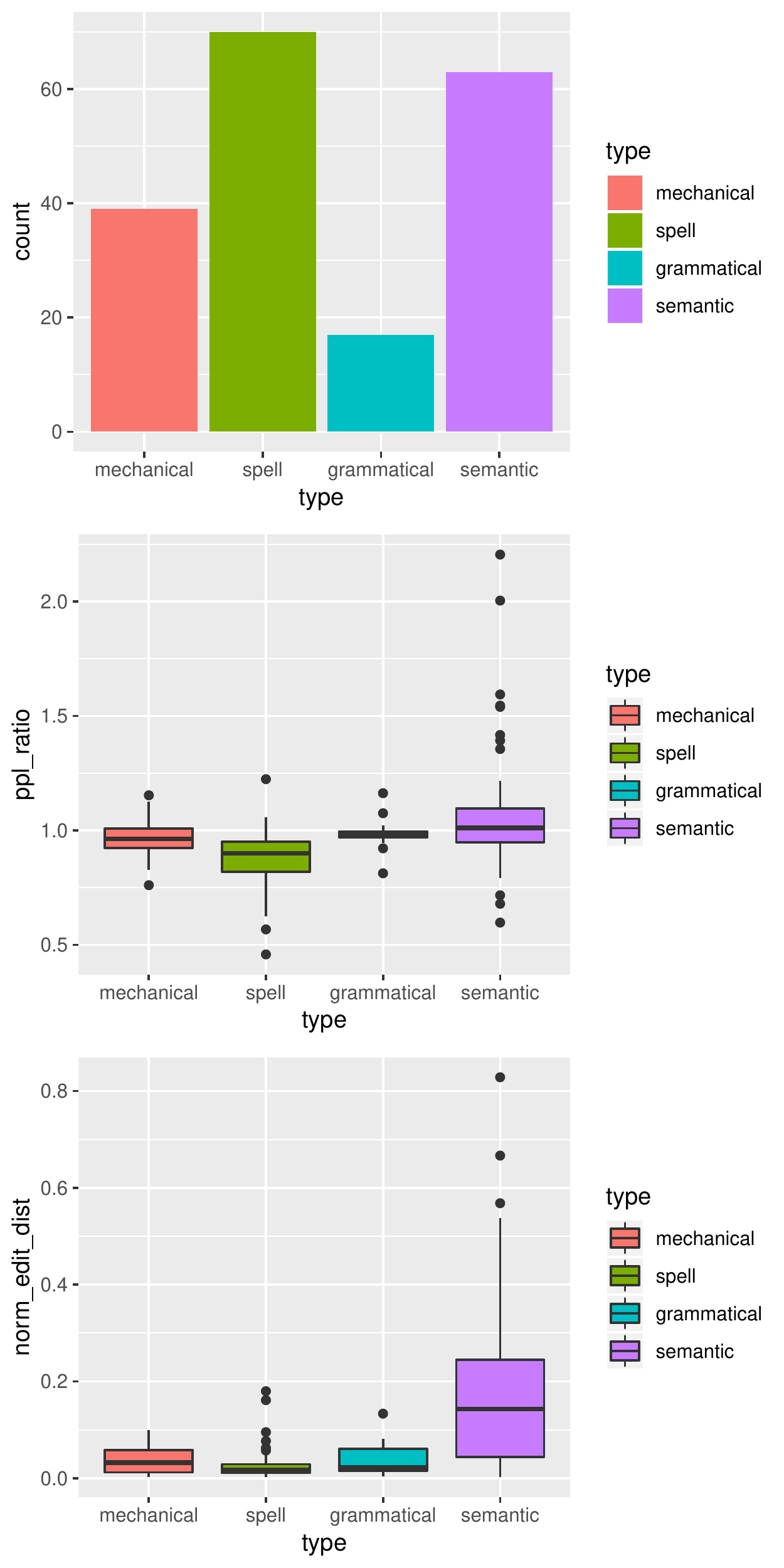}
\caption{Distribution of counts, perplexity ratio, and normalized edit distance per category}
\label{fig:plots}
\end{center}
\end{figure}

Finally, after annotating a small amount of samples for the three languages, we computed some basic statistics about each edit that may help in classifying typo edits from non-typo ones. Specifically, we computed three statistics:

\begin{enumerate}
    \item Ratio of the target perplexity over the source calculated by a language model
    \item Normalized edit distance between the source and the target
    \item Binary variable indicating whether the edit purely consists of changes in numbers
\end{enumerate}
The rationale behind the third feature is that we observed that purely numerical changes always end up being tagged as semantic edits.

The perplexity of a text ${\mathbf x} = x_1 x_2, ..., x_L$ is defined by:
\begin{equation}
    PP({\mathbf x}) = 2^{-H({\mathbf x})}, H({\mathbf x}) = \sum_{i} p(x_i) \log p(x_i),
\end{equation}
where $p(x)$ is determined by a trained language model. We hypothesize that perplexity captures the ``fluency'' of the input text to some degree, and by taking the ratio between the source and the target, the feature can represent the degree to which the fluency is improved before and after the edit.

As for the language model, we trained a character level Long Short Term Memory
(LSTM) language model developed in \cite{Merity:2018} per language, which consists of a trainable embedding layer, three layers of a stacked recurrent neural network, and a softmax classifier. 
The LSTM hidden state and word embedding sizes are set to be 1000 and 200, respectively.
We used 100,000 sentences from the W2C Web Corpus~\cite{Majlis2012LanguageRO} for training (except for Chinese, where we used 28,000 sentences) and 1,000 sentences for validation for all the languages.

The normalized edit distance between the source $\mathbf{x} = x_1 x_2, ..., x_{L_x}$ and the target $\mathbf{y} = y_1 y_2, ..., y_{L_y}$ is defined by:
\begin{equation}
    \tilde d({\mathbf x}, {\mathbf y}) = \frac{d({\mathbf x}, {\mathbf y})}{\max(L_x, L_y)},
\end{equation}
where $d({\mathbf x}, {\mathbf y})$ is the (unnormalized) edit distance between ${\mathbf x}$ and ${\mathbf y}$. This feature can capture the amount of the change made between the source and the target, based on our hypothesis that many typo edits only involve a small amount of changes.

See Figure \ref{fig:plots} for an overview of the distributions of these computed statistics per category for English. We observed similar trends for other two languages (Chinese and Japanese), except for a slightly larger number of spell edits, mainly due to the non-Latin character conversion errors. We also confirmed that the difference of perplexities between the source and the target for typo edits (i.e., mechanical, spell, and grammatical edits) was statistically significant for all three languages (two-tailed t-test, $p < .01$). This means that these edits, on average, turn the source text into a more fluent text in the target.

\subsection{Classification of Typo Edits}

\begin{table}[!t]
\begin{center}
\begin{tabular}{@{}lrrr@{}}
\toprule
Language & Precision & Recall & F1    \\ \midrule
English  & 0.874     & 0.969  & 0.917 \\
Chinese  & 0.872     & 0.930  & 0.896 \\
Japanese & 0.900     & 0.968  & 0.933 \\ \bottomrule
\end{tabular}
\caption{The cross validation result of typo edit classifiers}
\label{table:classification}
\end{center}
\end{table}

We then built a logistic regression classifier (with no regularization) per language using the annotated edits and their labels. The classifier has only three features mentioned above plus a bias term. We confirmed that, for every language, all the features are contributing to the prediction of typo edits controlling for other features in a statistically significant way $(p < .05)$. Table \ref{table:classification} shows the performance of the trained classifier based on 10-fold cross validation on the annotated data. The results show that for all the languages mentioned here, the classifier successfully classifies typo edits with an F1-value of approx. 0.9. This means that the harvested edits are fairly clean in the first place (only one third is semantic edits versus others) and it is straightforward to distinguish the two using a simple classifier. In the GitHub Typo Corpus, we annotate every edit in those three languages with the predicted ``typo-ness'' score (the prediction probability produced from the logistic regression classifier) as well as a binary label indicating whether the edit is predicted as a typo, which may help the users of the dataset determine which edits to use for their purposes.

\section{Analyses}

\begin{table*}[ht]
\begin{center}
\begin{tabular}{@{}lrrrr@{}}
\toprule
Language & \# commits & \# typo edits & \# all edits & \# chars    \\ \midrule
English          & 197,019   &  255,056  & 339,430 & 59,817,613 \\
Chinese (smpl.)  &   2,369   &    3,153  &   3,991 &    885,336 \\
Japanese         &   1,015   &    1,507  &   1,716 &    344,778 \\
Russian          &     837   &       --- &   1,600 &    509,887 \\
French           &     533   &       --- &   1,130 &    335,755 \\
German           &     419   &       --- &     700 &    215,948 \\
Portuguese       &     315   &       --- &     640 &    208,694 \\
Spanish          &     301   &       --- &     578 &    169,218 \\
Korean           &     206   &       --- &     442 &     89,852 \\
Hindi            &     158   &       --- &     197 &     34,277 \\
Others           &   1,760   &       --- &   2,631 &    571,068 \\ \midrule
Total            & 203,270   &  259,716  & 353,055 & 63,182,426 \\ \bottomrule
\end{tabular}
 \end{center}
\caption{Statistics of the dataset (top 10 languages)}
\label{table:stats}
\end{table*}

In this section, we provide detailed quantitative and qualitative analyses of the GitHub Typo Corpus.

\subsection{Statistics of the Dataset}

Table \ref{table:stats} shows the statistics of the GitHub Typo Corpus, broken down per language\footnote{Note that a commit is considered to be of a language if it contains at least one edit in that language; the commit numbers do not add up to the total.}. The distribution of languages is heavily skewed towards English, although we observe the dataset includes a diverse set of other languages. There are 15 languages that have 100 or more edits in the dataset.

In addition to an obvious fact that a large fraction of the code on GitHub is written in English, one reason of the bias towards English may be due to our commit collection process, where we used an English keyword ``typo'' to harvest eligible commit. Although it is a norm on GitHub (and in software development in general) to write commit messages in English no matter what language you are working in, we may be able to collect a more diverse set of commits if we build models to filter through commit messages written in other languages, which is future work.

\subsection{Distribution of Atomic Edits}

\begin{figure}[!t]
\begin{center}
\includegraphics[scale=0.5]{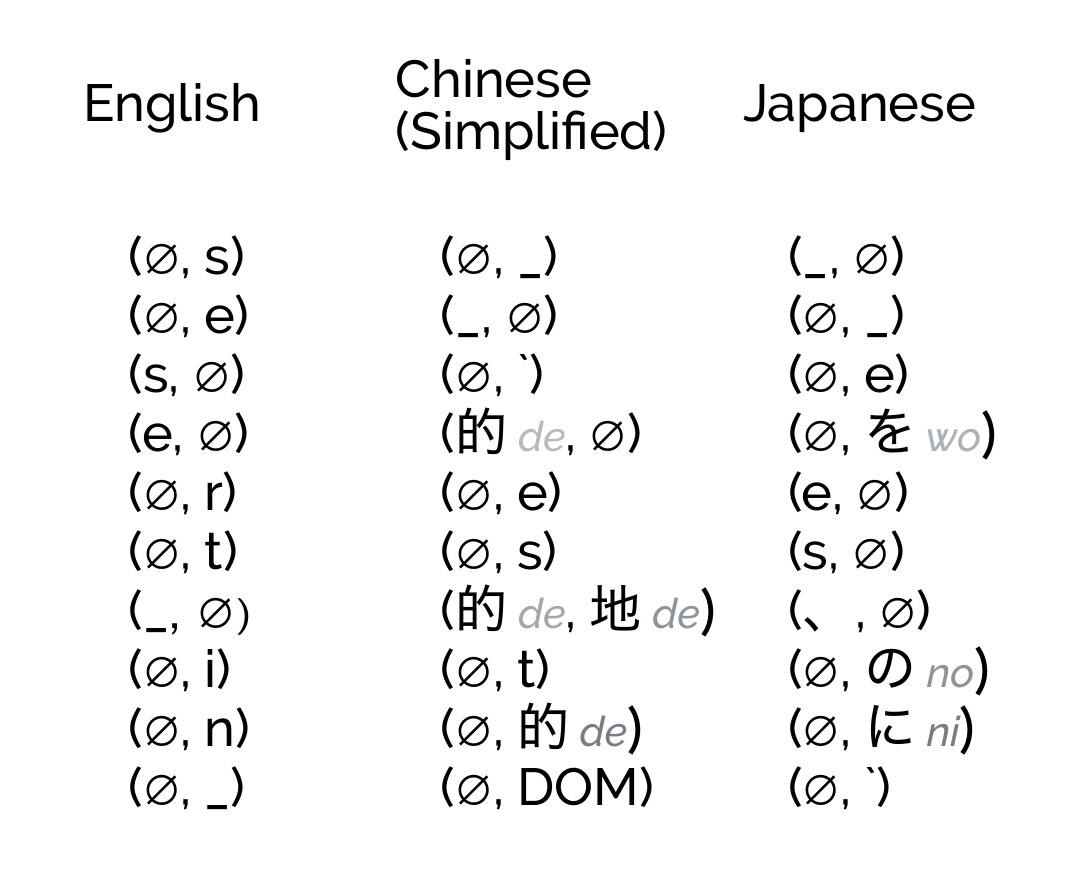}
\caption{Most frequent atomic edits per language. Underscore \_ corresponds to a whitespace and $\phi$ is an empty string.}
\label{fig:atomic_edits}
\end{center}
\end{figure}

In order to provide a more qualitative look into the dataset, we analyzed all the edits in the top three languages and extracted atomic edits. An atomic edit is defined as a sequence of contiguous characters that are inserted, deleted, or substituted between the source and the target. We extracted these atomic edits by aligning the characters between the source and the target by minimizing the edit distance, then by extracting contiguous edits that are insertion, deletion, or substitution.

As one can see from Figure \ref{fig:atomic_edits}, simple spelling edits such as inserting ``s'' and deleting ``e'' dominate the lists. In fact, many of the frequent atomic edits even in Chinese and Japanese are made against English words (see Figure \ref{fig:examples} for examples—you notice many English words such as ``GB-18030'' and ``Gemfile'' in non-English text). You also notice a number of grammatical edits in Chinese (e.g., confusion between the possessive particle {\it de} and the adjectival particle {\it de}) and Japanese (e.g., omissions of case particles such as {\it wo}, {\it no}, and {\it ni}). This demonstrates that the dataset can serve as a rich source of not only spelling but also naturally-occurring grammatical errors.

\begin{table*}[!ht]
\begin{center}
   \scalebox{0.85}[0.90]{ 
\begin{tabular}{lrrrrrrrr}
\toprule
\multicolumn{3}{c}{Edit type breakdown} &  \multicolumn{3}{c}{Aspell} & \multicolumn{3}{c}{Enchant} \\ \midrule
Type  & \# edits & \% total & Precision & Recall & F0.5  & Precision & Recall & F0.5 \\ \midrule
CONJ       & 1        & 0.7       & 1.000     & 0.000  & 0.000  & 1.000     & 0.000  & 0.000  \\
DET        & 5        & 3.5       & 1.000     & 0.000  & 0.000  & 1.000     & 0.000  & 0.000 \\
MORPH      & 3        & 2.1       & 0.000     & 0.000  & 0.000  & 0.000     & 0.000  & 0.000  \\
NOUN       & 10       & 7.0       & 0.000     & 0.000  & 0.000  & 0.091     & 0.100  & 0.093  \\
NOUN:INFL  & 2        & 1.4       & 0.000     & 0.000  & 0.000  & 0.000     & 0.000  & 0.000 \\
NOUN:NUM   & 1        & 0.7       & 1.000     & 0.000  & 0.000  & 1.000     & 0.000  & 0.000  \\
ORTH       & 15       & 10.5      & 0.118     & 0.133  & 0.121 & 0.067      & 0.133  & 0.074 \\
OTHER      & 16       & 11.2      & 0.000     & 0.000  & 0.000  & 0.000     & 0.000  & 0.000 \\
PREP       & 6        & 4.2       & 1.000     & 0.000  & 0.000 & 1.000     & 0.000   & 0.000  \\
PUNCT      & 16       & 11.2      & 1.000     & 0.000  & 0.000 & 0.000     & 0.000    & 0.000  \\
SPELL      & 56       & 39.4      & 0.563     & 0.643  & 0.577 & 0.500     & 0.625  & 0.521 \\
VERB       & 3        & 2.1       & 0.000     & 0.000  & 0.000  & 0.500     & 0.333    & 0.455 \\
VERB:FORM  & 3        & 2.1       & 1.000     & 0.000  & 0.000  & 1.000     & 0.000  & 0.000  \\
VERB:INFL  & 1        & 0.7       & 1.000     & 1.000  & 1.000 & 1.000      & 0.000  & 0.000  \\
VERB:SVA   & 2        & 1.4       & 1.000     & 0.000  & 0.000 & 1.000      & 0.000  & 0.000  \\
VERB:TENSE & 2        & 1.4       & 1.000     & 0.000  & 0.000 & 1.000      & 0.000  & 0.000 \\ \bottomrule
\end{tabular}
}
 \end{center}
 \caption{Distribution of edit types and the performance of spell checkers on the GitHub Typo Corpus}
\label{table:errant}
\end{table*}

\subsection{Evaluating Existing Spell Checker}

We conclude the analysis section by providing a comprehensive analysis on the types of spelling and grammatical edits, as well as the performance of existing spell checkers on the GitHub Typo Corpus. The first three columns of Table \ref{table:errant} show a breakdown of edit types in the aforementioned set of annotated typo edits in English (Section \ref{subsec:annotation}) analyzed by ERRANT~\cite{bryant:2017:automatic,felice:2016:automatic}. This shows that the dataset contains diverse types of edits, including orthographic, punctuation, and spelling errors. 

We then applied Aspell~\footnote{\url{http://aspell.net/}} and Enchant, two commonly used spell checking libraries, and measured their performance against each one of the edit types. The results show that the performance of the spell checkers is fairly low ($F0.5 \approx 0.5$) even for its main target category (SPELL), which suggests that the GitHub Typo Corpus contains many challenging typo edits that existing spell checkers may have a hard time dealing with, and the dataset may provide a rich, complementary source of spelling errors for developing better spell checkers and grammatical error correctors.

\section{Conclusion}

This paper describes the process where we built the GitHub Typo Corpus, a large-scale multilingual dataset of misspellings and grammatical errors along with their corrections harvested from GitHub, the largest platform for publishing and sharing git repositories. The dataset contains more than 350k edits and 64M characters in more than 15 languages, making it the largest dataset of misspellings to date. We automatically identified typo edits (be it mechanical, spell, or grammatical) versus semantic ones by building a simple logistic regression classifier with only three features which achieved 0.9 F1-measure. We provided detailed qualitative and quantitative analyses of the datasets, demonstrating that the dataset serves as a rich source of spelling and grammatical errors, and existing spell checkers can only achieve an F-measure of $\sim 0.5$.

We are planning on keep publishing new, extended versions of this dataset as new repositories and commits become available on GitHub. As mentioned before, collection of a more linguistically diverse set of commits and edits is also future work. We genuinely hope that this work can contribute to the development of the next generation of even more powerful spelling correction and grammatical error correction systems.

\section{Acknowledgements}

The authors would like to thank Tomoya Mizumoto at RIKEN AIP/Future Corporation and Kentaro Inui at RIKEN AIP/Tohoku University for their useful comments and discussion on this project.

\section{Bibliographical References}
\label{main:ref}
\bibliographystyle{lrec}
\bibliography{lrec2020W}


\end{document}